\documentclass[nohyperref]{article}

\usepackage{microtype}
\usepackage{graphicx}
\usepackage{subfigure}
\usepackage{booktabs} %
\usepackage{siunitx}

\usepackage{scalerel,xparse}

\usepackage{hyperref}

\usepackage{tabularx}

\usepackage{dsfont}
\usepackage{csquotes}
\newcommand{\norm}[1]{\left\lVert#1\right\rVert}
\usepackage[dvipsnames]{xcolor}

\usepackage[accepted]{icml2022}

\usepackage{amsmath}
\usepackage{amssymb}
\usepackage{mathtools}
\usepackage{amsthm}
\usepackage{amsfonts}
\usepackage{soul}

\usepackage[capitalize,noabbrev]{cleveref}

\theoremstyle{plain}
\newtheorem{theorem}{Theorem}[section]

\newtheorem{corollary}[theorem]{Corollary}
\theoremstyle{definition}
\newtheorem{definition}[theorem]{Definition}

\theoremstyle{remark}

\usepackage[textsize=tiny]{todonotes}

\def\mem{\text{mem}}

\renewcommand{\paragraph}[1]{\vspace{0.5em}\textbf{#1}}

\icmltitlerunning{Towards Differential Relational Privacy and its use in Question Answering}

\begin{document}
\twocolumn[
\icmltitle{Towards Differential Relational Privacy and its use in Question Answering}

\begin{icmlauthorlist}
\icmlauthor{Simone Bombari}{aws,ist}
\icmlauthor{Alessandro Achille}{aws}
\icmlauthor{Zijian Wang}{aws}
\icmlauthor{Yu-Xiang Wang}{aws}
\icmlauthor{Yusheng Xie}{aws}
\icmlauthor{Kunwar Yashraj Singh}{aws}
\icmlauthor{Srikar Appalaraju}{aws}
\icmlauthor{Vijay Mahadevan}{aws}
\icmlauthor{Stefano Soatto }{aws}
\end{icmlauthorlist}

\icmlaffiliation{aws}{AWS AI Labs}
\icmlaffiliation{ist}{IST Austria. Work conducted during an internship at AWS AI Labs}

\icmlcorrespondingauthor{Alessandro Achille}{aachille@amazon.com}

\icmlkeywords{Machine Learning, Differential Privacy, Relational Privacy, Question Answering, ICML}

\vskip 0.3in
]

\printAffiliationsAndNotice{}  %

\begin{abstract}

Memorization of the relation between entities in a dataset can lead to privacy issues when using a trained model for question answering. We introduce Relational Memorization (RM) to understand, quantify and control this phenomenon. While bounding general memorization can have detrimental effects on the performance of a trained model, bounding RM does not prevent effective learning. The difference is most pronounced when the data distribution is long-tailed, with many queries having only few training examples: Impeding general memorization prevents effective learning, while impeding only relational memorization still allows learning general properties of the underlying concepts.
We formalize the notion of Relational Privacy (RP) and, inspired by Differential Privacy (DP), we provide a possible definition of Differential Relational Privacy (DrP). These notions can be used to describe and compute bounds on the amount of RM in a trained model. We illustrate Relational Privacy concepts in experiments with  large-scale models for Question Answering.

\end{abstract}

\section{Introduction}\label{sec:1}

Consider training a deep neural network for question answering (QA) with data that may contain sensitive information, say a list of names and their corresponding phone numbers. The tendency of large neural networks to memorize training data can raise privacy concerns if sensitive information can be exfiltrated. 
Current tools from Differential Privacy (DP) help shield the data, but usually at the price of a steep performance loss, forcing users to either forgo privacy or accuracy. In some cases, however, the sensitive information may be {\em not} in the data itself, but rather in the {\em relation} among data points.  For example, the collection of phone numbers in a given area code may be public domain, but not their association with names. {\em Is it possible to extract utility from data while preserving privacy about relations within}? We call this variant of DP {\em Relational Privacy} (RP).

To frame RP more precisely and relate it to existing work in DP, we start with a simplified setting. Learning consists of estimating the parameters $w$ of a model using independent training samples $\{z_i\}_{i=1}^N$ that can be decomposed as  $z_i = (x_i, y_i)$, where $x_i$ contains information ({\em e.g.,} an email address or phone number) deemed private  {\em only if associated with} $y_i$, which denotes the remaining part of the sample ({\em e.g.,} a name or an identifier). This decomposition reflects the internal structure of the samples, which can be different depending on the task of interest and domain ({\em e.g.,} an image of an object and its background in a visual recognition system). Our first goal towards enabling RP is to {\em quantify and control the information about $x_i$ that an attacker can recover from the parameters of a trained model, $w$, and knowledge about the remaining part of the sample, $y_i$.}

The trade-off between accuracy and privacy in DP hardens when the task requires learning long-tailed distributions. In QA, for example, each of the  many possible queries is likely to have only few training examples; good performance of the system hinges on memorizing them. DP and its variants forbid learning {\em any} statistic of $x_i$ unless a large enough number of samples has been observed. We instead focus on enabling models to use $x_i$ so long as its relation to  $y_i$ is protected.

Previous work \cite{learningmem, whatmemorized} addressed \textit{memorization} as the amount of information a model captures from a single sample. In their setting, what matters more is the \textit{sensitivity} of a training algorithm with respect an individual training sample, which is the subject matter of DP. A short-cut to RP could be to declare a portion of the data private, for instance phone numbers in the example above, and train the model using Selective DP (sDP) \cite{shi2021selective}.  Imposing sDP constrains the information that can be extracted on the component $x_i$, which therefore bounds memorization of the relation between $x_i$ and $y_i$. While this can guarantee relational privacy, it also hampers the training process, for instance by preventing the model from learning the general format of phone numbers.

\subsubsection*{Marginal and Relational Memorization}

To study relational information, in \S\ref{sec:rm} we decompose the amount of information that a model can memorize about a sample $x_i$ in its parameters (weights) $w$ and that can be exfiltrated given knowledge of $y_i$ as the sum of two terms:
\begin{align}
    \text{mem}_W(X_i | Y_i)
    &= I(X_i ; W) -  I_\cap(X_i ; Y_i  ;W) \\
    &= \text{mem}^m_W(X_i) + \text{mem}^r_W(X_i, Y_i),
\end{align}
where $I$ denotes the mutual information. The first term measures the amount of information that the weights contain about $x_i$ {\em per se}, independent of its relation with other entities, ({\em e.g.,} which $x_i$'s were present in the training set, what are their statistics or distribution). We call this term \textit{Marginal Memorization} (MM); it is important to measure the risk of membership attacks, but it is not central to RP. The second, on the other hand,  measures the amount of information that the weights contain about {\em the relation between $x_i$ and $y_i$}. This is key to quantify the risk of adversarial extraction of private information about $x_i$ pertaining to a specific individual $y_i$ (or vice versa). We will call this term \textit{Relational Memorization} (RM).

\subsubsection*{Contributions}

Armed with the definition of Marginal Memorization and Relational Memorization, the next step is to derive estimates or bounds, so we can focus on controlling them. In particular, Relational Privacy hinges on controlling the latter.

We show that both DP and sDP only bound the sum of Relational and Marginal Memorization. We also show that Selective DP on the components $y_i$ alone bounds RM. We then state a necessary and sufficient condition that a training algorithm must respect in order to be Relational Private (RP), that is to completely avoid relational memorization (RM).

These are only first steps in the broad scope of Differential Relational Privacy (DrP), that aims to develop new tools to provide and guarantee privacy in the setting of this work. To further our investigation, we concentrate on a restricted class of models and present an analytical expression for memorization in linear models. This helps illustrate the concepts and may shed light on the more general problem. 

Finally, through experiments on Question Answering tasks, we efficiently quantify the amount of memorization in the trained models. In particular, we confirm that the emergence of the phenomenon can be associated to a particular instance of RM, which is a non-trivial result of the interaction between the particular internal structure of the data, the model used and the learning dynamics.

In summary:
\begin{enumerate}
    \item We introduce the concept of Relational Memorization to quantify the information that the weights of a trained model contain about the relation among parts of the data, rather than the raw data themselves. To the best of our knowledge, this area of investigation has received little attention despite the increasing importance of privacy in large deep neural networks;
    \item We quantify and characterize the amount of information that can be extracted from a trained model and part of a training sample in order to recover the remaining part. We establish connections with  Differential Privacy, and introduce the notion of Relational Privacy, with possible algorithmic constraints and methods to foster it -- which we refer to as  Differential Relational Privacy;
    \item We empirically probe some of the concepts introduced on Question Answering tasks. We provide an efficient policy to estimate memorization in this setting. We believe these to be the first experiments on memorization and privacy in a relational setting for QA. 
\end{enumerate}

\section{Related Work}

Differential Privacy \cite{DP} is an established tool to provide strong guarantees on the maximum amount of information that any attacker can extract from a trained model about an individual training sample. In particular, it allows users to select the desired trade-off between privacy and accuracy of the model, mediated by a privacy parameter $\varepsilon$, which can be chosen based on the application.
Specifically designed training algorithms, such as DP-SGD \cite{deeplearningDP}, allow training machine learning models while guaranteeing a given  $\varepsilon$-DP requirement. DP-SGD can be further improved by adapting to the structure of deep networks \cite{DPlowrank, adaptiveclipping}.

However, even with the current techniques,  DP is a strong constraint and training large machine learning models while guaranteeing strong privacy for each sample remains challenging \cite{DPneedsbetter}, since large models can easily memorize information about the training set \cite{attacks1, attacks2}. 
This is especially true on long-tailed tasks, where 
each class may only have few representatives in the training set. In this case, it may be difficult, or impossible, to separate the general information about the class from that of individual samples \cite{learningmem, whatmemorized}. Under these conditions, learning with  strong privacy constrain can severely impact accuracy.
To address this issue, \citet{shi2021selective} proposed Selective DP, that aims to improve utility in the case where only a subset of the information in each sample is considered private. Our work follows this generail aim, to  mitigate the utility loss from DP. In particular, we study the setting where it is not individual components of the samples that should be considered private, but rather their relation.

There is a strong relation between privacy and memorization in inductive learning. Memorization is often defined as the influence a training sample has on the model outputs \cite{learningmem, whatmemorized, uniqueinfo}.
Memorization has been studied in particular in Natural Language Processing (NLP), where it has been observed that large language models can output snippets of training text, when queried the right prefix \cite{carlini2021extracting, carlini2019secret}. We note that such large models are usually pre-trained on a reconstruction tasks, such as predicting the missing words in a sentence \cite{t5,GPT3}. This naturally encourages models to memorize the input verbatim, in order to achieve low training loss. Similarly, on downstream tasks like \textit{summarization} and \textit{translation}, where the correct ground truth target may be ambiguous, it has also been observed that memorized content may be outputted \cite{hallucinsummar, hallucintranslation}.

On the other hand, we focus on the setting where most of the privacy concerns are on the relation among data, such as in Question Answering (QA). In QA, a language model is trained to answer questions about an input text. In particular, QA datasets consists of samples organized as  a \textit{context}, a \textit{question}, and a ground-truth \textit{answer} to it. In principle, a well performing model \textit{does not need to memorize} to achieve zero training error, as the correct answer can be inferred from the context. 
However, we show that memorization is possible even in this setting. To the best of our knowledge, the empirical study of memorization in QA models is largely unexplored.

\section{Relational Memorization and Privacy}
\label{sec:rm}

We investigate the scenario where the training set is composed of $N$ independent training samples $S = \{z_i\}_{i=1}^N$, such that each $z_i$ can be decomposed as  $z_i = (x_i, y_i)$, where $x_i \in \mathcal X$ and $y_i \in \mathcal Y$. We use capitalized letters to refer to random variables, and lower-case for samples from their distribution. Let $\mathcal{A}$ be a stochastic training algorithm, which takes as input the dataset $S$ and outputs a probability distribution $p(w|S) = \mathcal A(S)$ over the weights space $\mathcal W$. We want to characterize the information being memorized in the weights $W$ about a sample $X_i$, and which can potentially be exfiltrated given the knowledge of $Y_i$:
\begin{equation}
\label{eq:mem-def}
    \text{mem}_W(X_i | Y_i) = \sup_\mathcal R I (\mathcal R(W, Y_i)  ; X_i) - I(Y_i  ; X_i)
\end{equation}
where $\mathcal R$ is a particular policy of recovery, that uses the trained model $W$ and the available part of the sample $Y_i$. The supremum of the mutual information between the recovery and the unknown attribute describes the success of the attack. The mutual information between the two attributes represents the information an attacker already has about $X_i$ given $Y_i$, without accessing to the model, which we subtract. For example, a name could already provides some information about the country of the individual, and therefore to the prefix of their phone number.

From the Data Processing Inequality and other basic properties of Mutual Information, we have:
\begin{equation}
    \begin{aligned}
    \text{mem}_W(X_i | Y_i) &= \sup_\mathcal R I (\mathcal R(W, Y_i) ; X_i) - I(Y_i  ; X_i)\\
    &= I((W, Y_i) ; X_i) - I(Y_i  ; X_i)\\
    &= I(W ; X_i) -  I_\cap(X_i ; Y_i  ;W) \\
    &=: \text{mem}^m_W(X_i) + \text{mem}^r_W(X_i, Y_i),
    \end{aligned}
\end{equation}
where we defined the Marginal Memorization (MM):
\begin{equation}
    \text{mem}^m_W(X_i) := I(W ; X_i),
\end{equation}
and the Relational Memorization (RM):
\begin{equation}
    \text{mem}^r_W(X_i, Y_i) := -  I_\cap(X_i ; Y_i ;W).
\end{equation}
Here $I_\cap(X ; Y ;Z) := I(X, Y) - I(X, Y|Z)$ is the interaction information between the three random variables, which can also be negative. $\text{mem}_W(X_i | Y_i)$ is always positive, since it can also be written as $\text{mem}_W(X_i | Y_i) = I(W;X_i | Y_i)$.

Marginal Memorization describes how much information the model contains about the random variable $X_i$, independently on its relation with $Y_i$. This quantity is equivalent to the information we would be able to extract from the model $W$ about $X_i$ without any knowledge of the context $Y_i$. Relational Memorization (RM) instead represents how knowledge of $Y_i$ could reveal information about the associated entity $X_i$. We are interested on the conditions under which an algorithm $\mathcal A$ can guarantee an upper bound on these two separate terms.

\subsection{Towards Differential Relational Privacy}

We now want to derive an algorithmic constraint that limits the amount of Relational Memorization. 
In particular, to guarantee that an attacker is unable to access additional information about the relation between $X_i$ and $Y_i$ if given access to the weights, we want $\mem^r_W(X_i, Y_i) = I(X_i; Y_i ) - I(X_i; Y_i | W) \leq 0$. \citet{mutualinfo} study under which conditions the mutual information between two random variables $I(X ; Y)$, is larger or smaller than their mutual information conditioned on a third random variable, $I(X; Y |W)$. 

Adapting their results to our setting, we get the following necessary and sufficient condition to guarantee that a model does not memorize any relation between entities in the data.

\begin{theorem}[\citet{mutualinfo}]
\label{th:corr}
Given a fixed algorithm $\mathcal A : (\mathcal X \times \mathcal Y)^n \to \mathcal W$, if, for every joint distribution $P_{XY}$ such that the independent samples $(X_i, Y_i) \sim P_{XY}$, we have $\textup{mem}^r_W(X_i, Y_i) \leq 0$, then the model does not fit spurious correlations between $Y_i$ and $X_i$, which means that
\begin{equation}\label{eq:factorizable}
    p(w | x_i, y_i) = r(w, x_i) s(w, y_i),
\end{equation}
where $r, s$ are generic functions from $\mathcal W \times \mathcal X$ to $\mathbb R$ and from $\mathcal W \times \mathcal Y$ to $\mathbb R$.

Vice versa, if the algorithm $\mathcal A$ generates a conditional distribution of $W$ that factorizes with respect $x_i$ and $y_i$, then
\begin{equation}
    \textup{mem}^r_W(X_i, Y_i) \leq 0.
\end{equation}
\end{theorem}

The previous theorem provides necessary conditions that a training algorithm should satisfy in order to avoid incurring in Relational Memorization. In particular, a model does not memorize the relation between $X_i$ and $Y_i$, for every $i$, if and only if it satisfies the following definition.

\begin{definition}[Factorized model]
\label{def:factorized-model}
We call a model $W \sim A(S)$ \textit{factorized} if, for every $i$, there exist two functions $r(w, x_i)$, $s(w, y_i)$ such that:
\[p(w | x_i, y_i) = r(w, x_i) s(w, y_i),\]
for all $(x_i, y_i) \in \mathcal{X} \times \mathcal{Y}$.
\end{definition}

Equation \ref{eq:mem-def} allows us to measure, \textit{after} training, whether a given model memorizes a sample. However, in practice it is important to design algorithms that can guarantee, \textit{before} training, that relational information will not be memorized. To this end, we introduce the following definition.

\begin{definition}[Differential Relational Privacy]
\label{def:drp}
An algorithm $\mathcal A: \mathcal S \to \mathcal W$ is $\varepsilon$-Differential Relational Private (DrP) with respect to $x$ and $y$ 
if there exists a random variable $\hat{W}$ factorized with respect to $X_i$ and $Y_i$, such that, for all $w$,
\begin{equation}
    p(w | x_i, y_i) \leq e^\varepsilon \hat{p}(w | x_i, y_i),
\end{equation}
where $p(w|x_i, y_i)$ and $\hat{p}(w|x_i, y_i)$ are the conditional probability distributions of $W$ and $W'$ respectively.
\end{definition}
This definition encodes the requirement of having a model \enquote{$\varepsilon$-close} to a factorized one.
The previous considerations and results suggest that encouraging disentanglement and penalizing correlations promote Relational Privacy. This can be achieved, for example, using data augmentation, which is already known to provide increasing privacy guarantees \cite{dataaug}.

\subsection{Connection with Differential Privacy}
\label{sec:rm-and-dp}

Differential Privacy (DP) \citep{dwork2006calibrating} is a formal definition of privacy that limits the usage of information from a sample when training a model. In this section, we show how different notions of DP protect against different types of memorization. We start by recalling the definition of DP.
\begin{definition}
An algorithm $\mathcal A: \mathcal S \to \mathcal W$ is $\varepsilon$-DP if, for any two adjacent training sets $S, S' \in \mathcal{S}$, and for every $w \in \mathcal{W}$, the following holds.
\begin{equation}
    p (w | S) \leq e ^ \varepsilon p( w | S'),
\end{equation}
where adjacent means that $S$ and $S'$ differ only by one sample $s$.
\end{definition}
As specified before, $p(w|S) = \mathcal A(S)$ is the probability distribution of the weights after training on the dataset $S$ using a (stochastic) training algorithm $\mathcal{A}$.

DP requires that each individual sample by itself cannot have a large effect on the final outcome of the training. This may make learning difficult in few-shot or long-tail tasks, where only a few samples are available for a given class or question. 
Several alternative definitions of DP have been proposed trying to gain better utility on a dataset by restricting privacy to a subset of the information in a sample. In particular, we refer to \citet{shi2021selective}, that defines Selective DP (sDP) for structured data.
Their definition, in our settings (if we consider $x$ to be the private component), takes this form:
\begin{definition}
An algorithm $\mathcal A: \mathcal S \to \mathcal W$ is $\varepsilon$-sDP if, for any two selective-adjacent training sets $S, S' \in \mathcal{S}$, and for every $w \in \mathcal{W}$, the following holds.
\begin{equation}
        p (w | S) \leq e ^ \varepsilon p( w | S'),
\end{equation}
where selective-adjacent means that $S$ and $S'$ differ only over the private component of one sample. In other words, only one sample is different in the two datasets, and this sample will take the forms $s = (x, y)$ and $s' = (x', y)$.
\end{definition}

Notice that strict sDP or DP conditions with $\varepsilon = 0$ would imply a factorized model (Definition~\ref{def:factorized-model}) with $r = 1$ for sDP and complete independence in DP.

We now show that, for any $\varepsilon$, Selective DP (and consequently also the stronger DP) are also bound on the maximum possible amount of memorization in the model (proof in Appendix~\ref{app:1}, based on \citet{dpinfo}):
\begin{theorem}\label{th:selectmem}
An $\varepsilon$-sDP Algorithm (over the $x$ component) guarantees, for all $i$-s,
\begin{equation}
    \textup{mem}_W(X_i | Y_i) \leq \varepsilon.
\end{equation}
\end{theorem}
\begin{corollary}
An $\varepsilon$-DP Algorithm guarantees, for all $i$-s,
\begin{equation}
    \textup{mem}_W(X_i | Y_i) \leq \varepsilon.
\end{equation}
\end{corollary}

Since we are interested in bounding the Relational Memorization our model has with respect to $x_i$ and $y_i$, we provide the following bound (proof in Appendix \ref{app:2}):
\begin{theorem}\label{th:selectmemy}
An $\varepsilon$-sDP Algorithm (over the $y$ component) guarantees, for all $i$-s,
\begin{equation}
    \textup{mem}^r_W(X_i, Y_i) \leq \varepsilon.
\end{equation}
\end{theorem}

The intuition behind Theorem \ref{th:selectmemy} is that, if we want to bound the information the model has about the relation between $x_i$ and $y_i$, bounding the information the model has about $y_i$ is sufficient. The previous definitions and results show that DP is sufficient to bound Relational Memorization. The theorem also suggests that moving the sDP constraint over the \textit{non-sensitive} component of the sample can be a solution (for example, a phone number alone, without the name of the owner, may not present privacy issues), if it impacts positively the performances of the model.

\section{Empirical Measurement of Memorization}\label{subsec:memmeas}

Directly estimating the Relational Memorization score in Equation~\ref{eq:mem-def} on real datasets and models may prove challenging.  Here, we propose an efficient way to estimate the expected RM, based on performing inference over sub-parts of training and validation samples.

Consider a dataset $S = \{(x_i, y_i)\}_{i=1}^N$, and let $x_i$ denote the target information to extract, while $y_i$ denotes the remaining part of the sample.
For example, in an image classification task, $y_i$ would represent the background, and $x_i$ the right label of the object that was corrupted from the sample image $z_i$. In Question Answering, $y_i$ would be the context of sample $z_i$, after we removed the sub-strings that enable to give the right answer ($x_i$) to the relative question (see Table \ref{tab:examples}, for example).

We will work under the following assumptions:
\begin{enumerate}
    \item the training algorithm is symmetric with respect to the input samples, which leads to $I(W, Y_i ; X_i) = I(W, Y_j ; X_j)$, for every couple of training samples. This term will be indicated by $I(W, Y ; X)$. This assumption is satisfied, for example, by stochastic gradient descent (SGD).
    \item $p(x_i) \simeq c = e^{-H(X)}$. We are assuming ground truth to be equi-probable. This may be the case, for example, in image classification, and is a reasonable approximation on the numerical answers on the QA datasets we will use in the experiments;
    \item an almost optimal policy of attack is using corrupted samples for inference (this also resembles what we do with linear models in \S\ref{subsec:linear}). This permits us to write $p(x_i | y_i, w) = q_w((-, y_i))_{x_i}$, where $q_w$ is the probability distribution described by the trained model. This assumption stands close to the setting that considers black-box attacks on the model;
    \item either the model has strong confidence on the reconstruction, which means $q_w((-, y_i))_{x_i} \simeq 1$, or almost no confidence, $q_w((-, y_i))_{x_i} \simeq p(x_i) \simeq c$.
    \item since the algorithm is symmetric, $I(X ; W) \simeq I(X; \mathcal R(W))$, where $\mathcal R$ is an attack that interrogates the model in a way that each sample $x_i$ is returned with probability $1/n$;
    \item we assume $nc \gg 1$. Or alternatively, $\log n \gg H(X)$. This, again, is reasonable in our experimental settings.
\end{enumerate}

We say that an attack over the sample $i$ is successful if $\text{argmax } q_w((-, y_i)) = x_i$ (the model returns the right label or answer given the corrupted input). If we call $r_{tr}$ and $r_{val}$ the success rates of the attacks on the training and validation set, it can be shown (see Appendix \ref{app:4}) that the expected RM the model has for $x$ and $y$ takes the form
\begin{equation}
    \text{mem}^r = H(X) (r_{tr} - r_{val}).
\end{equation}
The expected Relational Memorization between the two components, on our dataset, will be the difference of the recovery scores between the training set and a validation set.

In practice, in our experiments, we will have small control on the entropy of the random variable $X$. Thus, we will compute a normalized version of the RM,
\begin{equation}\label{eq:empmem}
    \text{m}^r = r_{tr} - r_{val}.
\end{equation}

\begin{table*}[ht]

\begin{center}
\begin{small}
\begin{tabular}{p{5.5cm} p{4.5cm} p{2.5cm} p{1.25cm} p{1.4cm}}
\toprule
Context & Corrupted Context & Question & Answer & Answer on Corrupted\\
\midrule
On February \textcolor{red}{6}, \textcolor{red}{2016}, one day before her performance at the Super Bowl, Beyoncé released a new single exclusively on music streaming service Tidal called "Formation". &
On February , , one day before her performance at the Super Bowl, Beyoncé released a new single exclusively on music streaming service Tidal called "Formation". &
What day did Beyonce release her single, Formation? &
February 6, 2016 & 
February \\

\midrule
In the county, the population was spread out with \textcolor{red}{23}.\textcolor{red}{5}\% under the age of \textcolor{red}{18}, \textcolor{red}{7}.\textcolor{red}{8}\% from \textcolor{red}{18} to \textcolor{red}{24}, \textcolor{red}{28}.\textcolor{red}{5}\% from \textcolor{red}{25} to \textcolor{red}{44}, \textcolor{red}{25}.\textcolor{red}{9}\% from \textcolor{red}{45} to \textcolor{red}{64}, and \textcolor{red}{14}.\textcolor{red}{2}\% who were \textcolor{red}{65} years of age or older.  The median age was \textcolor{red}{40} years. For every \textcolor{red}{100} females, there were \textcolor{red}{93}.\textcolor{red}{8} males.  For every \textcolor{red}{100} females age \textcolor{red}{18} and over, there were \textcolor{red}{90}.\textcolor{red}{5} males.  & 
In the county, the population was spread out with .\% under the age of , .\% from  to , .\% from  to , .\% from  to , and .\% who were  years of age or older.  The median age was  years. For every  females, there were . males.  For every  females age  and over, there were . males. &
How many in percent wasn't under 18 for the county?  &
76.5 &
76.5\\
\bottomrule
\end{tabular}
\end{small}
\end{center}
\vskip -0.1in
\caption{\textbf{Examples from the SQuAD and DROP training sets.} The model may give the correct answer on corrupted samples, even though there is not the necessary information in the context anymore (second example). This happens because the model memorizes the relation between the context and the ground truth answer. If the document contains private information, the model may still output that information even though that information is not in the query.}\label{tab:examples}

\end{table*}

\section{Relational Memorization in Linear Models}\label{subsec:linear}

\begin{figure}
\vskip 0.2in
\begin{center}
\centerline{\includegraphics[width=\columnwidth]{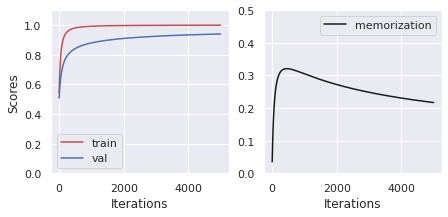}}
\caption{\textbf{Memorization can decrease over time.} While in over-parametrized models memorization generally increase over time,  we show that when the size of the linear model in \S \ref{subsec:linear} is comparable with the number of samples (here $d = n = 1000$) gradient descent may exhibit a non-trivial dynamics where the model first memorizes relations, but forget them later on \textbf{(right)}.
}
\label{fig:synthetic}
\end{center}
\vskip -0.2in
\end{figure}

In order to better gauge what affects Relational Memorization, we explicitly compute  Relational Memorization in linear models on a prototypical regression task. Deep learning models have been shown to mimic linear models in the limit of wide networks \cite{ntk, nnlinear}, and linear approximations of them have been shown to perform comparably to their non-linear counterparts in  different tasks \cite{uniqueinfo, LQF}. Therefore, studying this setting can provide insights for more general models.

Let us suppose we have a dataset with $N$ samples $(z_i, g_i)_{1 \leq i \leq N}$, where $g_i$ is the ground truth of the inputs $z_i$'s. Let us consider the case where $z_i = (x_i, y_i)$ and $g_i = x_i$. In particular the inputs are vectors made by two independent components, $x_i \in \mathbb{R}$ and $y_i \in \mathbb{R}^d$. The labels are a function of only the first component of $z_i$. We define $Z$ to be the matrix with $z_i$ as the $i$-th row, and at the same way $X$ and $Y$. In particular, we have $Z = [X, Y]$. As before, we can think about $x_i$ as the component of the samples we want to recover through $y_i$. We will study how the relation between $d$ and $n$ can influence the learned model.

The empirical loss can be written as
\begin{equation}\label{eq:loss}
    L(w) = \frac{1}{2n} \sum_i (w ^\top z_i - x_i)^2.
\end{equation}
The minimizer $w^* = \arg \min L(w)$ satisfies the following equation
\begin{equation}
    Z ^\top (Zw^* - X) = 0.
\end{equation}

When $d < n$, the solution is unique (if the samples are not linearly dependent) and it is easy to verify that $w^* = e^{d + 1}_1$, where $e^p_q$ is the $q$-th element of the canonical base in $\mathbb R^p$. This means that neither the $x_i$-s nor the $y_i$-s influence the model, then Relational Memorization is $0$. %

When $d \geq n$ the solution is no longer unique and different algorithms may converge to different global minima, which may exhibit different memorization. In Appendix~\ref{app:3}, we study memorization when training with gradient descent, starting with zero initialization $w(0) = 0$. The main results are that, after training, feeding the model the query $\tilde z_i = [0, y_i]$, an attacker would be able to recover $x_i$. In other words, $y_i$ behaves as a \textit{decryption key} to obtain $x_i$ through the model. Moreover, in the limit of $d \gg n$, the solution becomes almost independent of the first components $x_i$, since overfitting dominates (see Appendix \ref{app:3}). Therefore, we have $\text{mem}^r_w(X_i, Y_i) \simeq \text{mem}_w(X_i | Y_i) = H(X)$.

These results suggest that,  if the \textit{nuisance} component of the samples is high dimensional, it will be easier to over-fit. In particular, over-parametrized models are more susceptible to Relational Memorization between the two parts of the sample. This can hold in more general contexts. For example, in a QA task, if the context contains the same string of the answer, even if with a different meaning, it can be over-fitted by the model, and \textit{related} to the ground truth answer to the question.

Finally, when $d \approx n$ so that the number of training samples is comparable with the size of the model, we observe non-trivial memorization dynamics even in simple linear models. In particular, in Fig.~\ref{fig:synthetic} we observe that some relations are first learned but then forgotten with additional training.

\section{Experiments}\label{sec:exp}

\paragraph{Datasets.} We empirically investigate Memorization in Question Answering (QA) tasks. We focus on two QA datasets: SQuAD \cite{squad} and DROP \cite{drop}. In these datasets, every sample is formed by a \textit{context}, a \textit{question}, and a ground-truth \textit{answer}. An example for both SQuAD and DROP, is provided in Table \ref{tab:examples}, respectively in the first and second row. DROP, differently from SQuAD, focuses its questions on discrete reasoning over paragraphs. Sometimes, this implies the absence of the ground-truth answer in the context, making it more difficult for language models to achieve high performance. In SQuAD, instead, the answer string is always contained in the context. Both are well-studied QA datasets, with a reasonably high fraction of numerical questions. SQuAD contains $22\%$ of questions where the answer contains a digit, while DROP contains $69\%$ of numerical questions. This allows an especially simple experimental design for the study of memorization. 

\paragraph{Models.} We make use of T5-small \cite{t5} and numnet+ \cite{numnet} to perform our experiments on the datasets mentioned above. We made use of the same hyperparameters listed in the relative references, to make our experiments easy to reproduce.

\subsection{Dataset Memorization}

To measure the amount of memorization in our trained models, we apply the policy described in \S \ref{subsec:memmeas}, applying directly Equation~\ref{eq:empmem}. Given the availability of numerical answers $x_i$ in the two datasets, we define $y_i$ to be equal to the sample $z_i$, after we remove all the digits from the context (we leave the question untouched), we call this procedure \textit{corruption}. We opted for this corruption procedure since it is fast and easy to implement.

There will be samples where it is still possible to obtain the right answer from the corrupted context. This is the case for non-numerical questions, for example. We mitigate this problem in SQuAD, where most of the answers are not numerical, measuring the RM only on its numerical subset.

Some numerical questions may still be solvable, since the answer is present in words in the context, or because of prior information in the model (T5 is pre-trained on web pages, where both datasets come from). Also, the model can of course return the right answer by chance. However, these factors are all considered in our definition of Memorization, when we subtract the term $I(X_i, Y_i)$.

We measure $\mem^r$ between $X_i$ and $Y_i$, to show that QA models can memorize. To compute the rate $r$ of success of the reconstruction attacks, we make use of both metrics F1 and EM (exact match). In the datasets we use, those metrics are computed asymmetrically between the training set and the validation set: in fact, on the validation set there are more possible ground truth answers, that are compared with the answer provided by the trained model. This biases the scores on the validation set to be slightly higher.

In the following paragraphs, we present our empirical findings, connecting the role that data, model, and training time have on the emergence of RM.

\begin{table}[t]
\label{sample-table}
\begin{center}
\begin{small}
\begin{sc}
\begin{tabular}{lS[table-format=-2.1]S[table-format=-2.1]|S[table-format=-2.1]S[table-format=-2.1]}
\toprule
 & \multicolumn{2}{c|}{T5/SQuAD} & \multicolumn{2}{c}{numnet+/DROP}  \\
\midrule
 & \multicolumn{1}{r}{EM} & \multicolumn{1}{r|}{F1} & \multicolumn{1}{r}{EM} & \multicolumn{1}{r}{F1$~~$} \\
\midrule
$s_{\text{tr}}$        & 79.2    &   89.0  & 79.9   &   84.7  \\
$s_{\text{val}}$       & 84.0    &   90.0  & 78.6   &   82.3 \\
$r_{\text{tr}}$        & 0.1     &   23.5  & 35.8   &   39.2 \\
$r_{\text{val}}$       & 1.9     &   26.0  & 31.5   &   34.7 \\
$m$                    & -1.8    &   -2.5  & 4.2    &   4.5 \\
\bottomrule
\end{tabular}
\end{sc}
\end{small}
\end{center}
\caption{\textbf{Memorization scores on SQUAD and DROP.} We report scores ($s_{\text{tr}}$, $s_{\text{val}}$), recovery scores ($r_{\text{tr}}$, $r_{\text{val}}$), and normalized memorization ($m$) for T5-small trained on SQuAD and for numnet+ trained on DROP. On SQuAD the scores computed only on the subsets of the datasets containing only numerical answers. The Relational Memorization results negative in SQuAD. This is caused by an asymmetry between the scoring metric on training and validation set, that provides more available correct answers on the validation set. These results suggest lack of memorization for SQuAD, but a meaningful amount of memorization for DROP.}\label{tab:numbers}
\vspace{-1em}
\end{table}

\begin{figure}[ht]
    \vskip 0.2in
    \begin{center}
    \centerline{\includegraphics[width=\columnwidth]{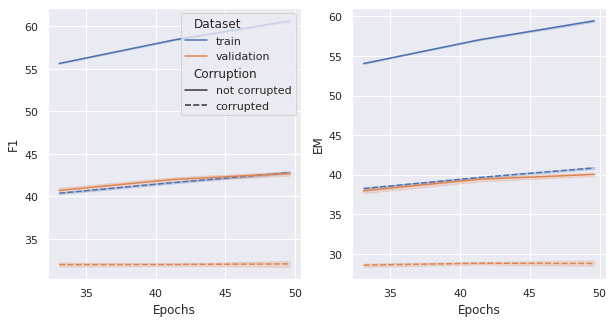}}
    \caption{\textbf{Complex QA datasets display memorization.} We report the F1 (left) and EM (right) scores when training a T5-small model on DROP, as a function of the training epoch. We observe that the a large gap between the corrupted train (dashed blue line) and the corrupted val (dashed orange) suggesting an high-memorization. In particular, in this range of epochs, we have on average $m_{\text{EM}} = 10.9$ and $m_{\text{F1}} = 9.6$.
    }\label{fig:t5drop}
    \label{fig:2}
    \end{center}
\end{figure}

\paragraph{Data.}
Using the T5-small model, and fine-tuning it over the SQuAD and DROP tasks, we can make the first empirical observations: on models that overfit, it is easier to witness the emergence of memorization. In SQuAD, there is little overfitting, and the answer is always explicitly contained in the context. In DROP, the answer is usually not explicitly contained in the context. However, the answer string is sometimes contained somewhere else, either in the context or in the question (see, for example, Table \ref{tab:examples}). At training time, the model overfits this string, \textit{relating} it to the ground truth answer. This results in the model being able to correctly reply to answers on the training samples, even if the informative part of the context has been removed. The results are showed in Table \ref{tab:numbers} and Figure \ref{fig:t5drop}.

\begin{figure*}[!ht]
\begin{center}
\centerline{\includegraphics[width=\textwidth]{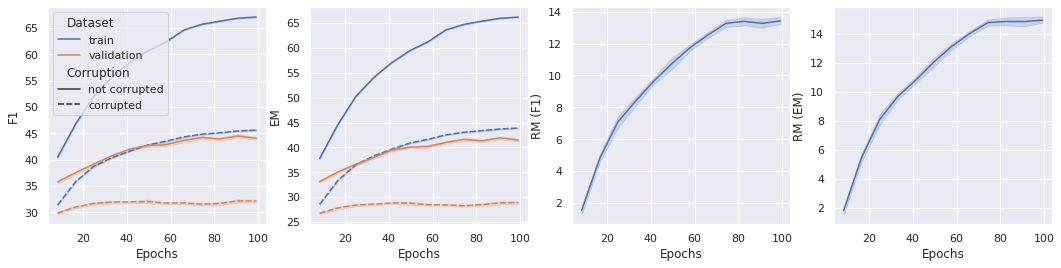}}
\caption{\textbf{Over-training stimulates memorization.} We show how longer training in an over-parametrized model affects Relational Memorization using T5-small on DROP. In this case, overtraining leads to increasing memorization (right two plots).}
\label{fig:training-time}
\end{center}
\vspace{-1em}
\end{figure*}

\paragraph{Model.} 
numnet+ \cite{numnet} is an architecture explicitly designed to reach good performance on the DROP dataset. Using a model with higher performances, overfitting decreases. However, we can observe that Memorization still occurs (see Table \ref{tab:numbers}). On T5-small, the extent of memorization is still larger ($10.9$ against $4.2$, in the EM metric), especially at larger epochs (Figure \ref{fig:2} and Figure \ref{fig:training-time}). This experiment shows how even high performing models can still exhibit memorization in QA tasks, but confirms that smaller generalization error usually implies lower memorization. Experiments on T5-base and findings on how the size of the model can affect memorization can be found in Appendix \ref{app:b}.

\paragraph{Training Time.} 
Overtraining is associated with memorization. In Figure \ref{fig:training-time}, we show that this is what we empirically see in training dynamics like T5-small over DROP. However, we see that memorization starts from epoch 10, while the validation accuracy keeps increasing until epoch 50. This implies that memorization cannot be associated with only \textit{post-learning} times.

This dynamics of memorization and learning can also behave very differently. For example, in the linear model discussed in \S\ref{subsec:linear}, if the number of parameters is larger but with the same order of magnitude of the data points, we observe a local maximum in the memorization instead (see Figure \ref{fig:synthetic}): the memorization first increases, but then decreases. This indicates that relational information is being forgotten as the training progresses, in contrast with the previous case, and with the usual empirical observation in over-parametrized models, where Memorization is directly associated with over-training (see Appendix \ref{app:3} for more details and an explanation of this phenomenon).

\section{Conclusions}

We study an under-explored privacy problem in trained machine learning models, where the concern is not the privacy of the data themselves, but of the \emph{relation} among their components. In particular, we decompose the memorization often observed in large machine learning models in different factors that can be studied and controlled independently. Our analysis allows extracting utility from the data while preserving the privacy of their relation. This is especially important for long-tail problems such as Question Answering, where each question may come with relatively few samples. We establish connection with Differential Privacy, showing that different notions of DP can bound different aspects of memorization. While memorization for large transformer models has been shown empirically on reconstruction tasks, such as next word prediction, we empirically study memorization in QA tasks which, in principle, do not require memorization to reach perfect training loss. Here we show that transformers can indeed still memorize sensitive information, but whether that happens depends on a combination of non-trivial factors.

In \S\ref{sec:rm-and-dp} we showed that Relational Memorization can be controlled through existing notions in differential privacy. However, these may be a blunt upper-bound to RM. Designing differential privacy algorithms that can take full advantage of the RM setting may further improve the accuracy of private model on long-tail data. Our Definition~\ref{def:drp} is a first step in this direction.

In addition, RM could go beyond Question Answering. QA problems provide a clear separation between private data and context. This simplifies the experimental design and the measurement of RM. However, the notion of RM applies more widely to other tasks, such as relations between entities in images. We leave extending RM to these compelling cases to future work.

\bibliography{bibliography}
\bibliographystyle{icml2022}

\newpage
\appendix
\onecolumn

\section{Additional Experiments}

\subsection{Memorization in T5-base on the DROP dataset}\label{app:b}

\begin{figure*}[ht]
\vskip 0.2in
\begin{center}
\centerline{\includegraphics[width=\textwidth]{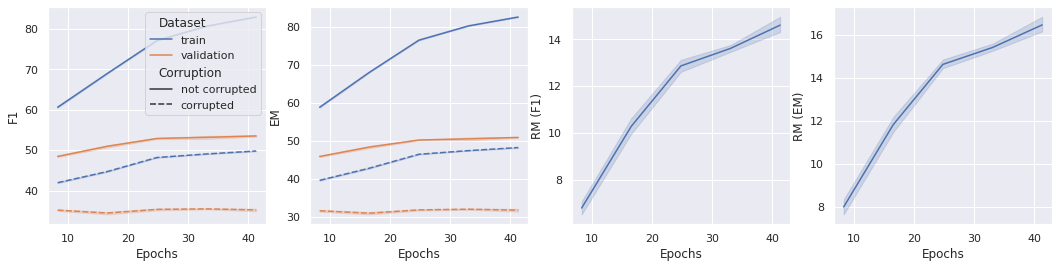}}
\caption{Memorization in T5-base on DROP. Performances, as well as Memorization, are higher in this model.}
\label{fig:t5base}
\end{center}
\vskip -0.2in
\end{figure*}

We perform the same experiments over the DROP task with the model T5-base, a bigger version of T5-small. From Figure \ref{fig:t5base}, we see that the value of RM at epoch 40 is already larger than RM at epoch 100 in the case of T5-small (Figure \ref{fig:t5drop}). Also the performances increase.

\section{Additional Proofs}

\subsection{Proof of Theorem \ref{th:selectmem}}\label{app:1}
\begin{proof}
Dropping the index $i$ to ease the notation
\begin{equation}\label{eq:firststep_app1}
    \text{mem}_W(X | Y) = I(W; X | Y) = 
     \int p(y) \int p(w, x | y) \log \frac{p(w, x | y)}{p(w | y) p (x | y)} = 
     \int p(w, x, y) \log \frac{p(w|x,y)}{p(w|y)}.
\end{equation}
From the definition of $\varepsilon$-Selective DP over $x$ we have, for all $x'$
\begin{equation}
    \begin{aligned}
    p(w|x, y) &\leq e^\varepsilon p (w|x', y) \\
    e^{-\varepsilon} &\leq \frac{p(w|x',y)}{p(w|x, y)}
    \end{aligned}
\end{equation}
Integrating both sides over $p(x' | y)$ gives
\begin{equation}
    e^{-\varepsilon} \leq \int_{x'}\frac{p(w|x',y)}{p(w|x, y)}p(x'|y) = \frac{p(w|y)}{p(w|x,y)}.
\end{equation}
Merging the last inequality, with equation \ref{eq:firststep_app1}, gives
\begin{equation}
    \text{mem}_W(X | Y) = I(W; X | Y) = \int p(w, x, y) \log \frac{p(w|x,y)}{p(w|y)} \leq \int p(w, x, y) \log e^\varepsilon = \varepsilon.
\end{equation}
\end{proof}

\subsection{Proof of Theorem \ref{th:selectmemy}}\label{app:2}
\begin{proof}
Dropping the index as in the previous proof, we can write
\begin{equation}
    \textrm{mem}^r_W(X, Y) = - I_\cap  (X_i ; Y_i  ;W) = I(W; Y | X) - I(W;  Y) \leq I(W; Y | X)
\end{equation}
In the previous proof we showed that $\varepsilon$-Selective DP over $x$ implies $I(W; X | Y) \leq \varepsilon$. In the same way, we can then prove that $\varepsilon$-Selective DP over $y$ implies $I(W; Y | X) \leq \varepsilon$, which proves the theorem.
\end{proof}

\subsection{Computation for RM in linear models, with $d \geq n$}\label{app:3}

Let the singular value decomposition of $Y$ be $Y = V D U^\top$, with $D$ a $n \times d$ matrix, with the singular values of $Y$. The solution of the optimization problem respects the following equation
\begin{equation}\label{eq:minim}
    Zw^* - X = 0.
\end{equation}
Also, since we assume $w(t=0) = 0$, we have that if $v \in \text{Ker}(Z)$, then $v^\top w(t) = 0$ for all $t$-s (the dynamics happens only outside this subspace). From Equation \ref{eq:minim}, we know that we can write the minimizer as
\begin{equation}\label{eq:v}
    w^* = e^{d + 1}_1 + \frac{v}{\norm{v}^2},
\end{equation}
where $v = [1, Y_r^{-1} X]$. $Y_r^{-1}$ is a right inverse of $Y$. It can be shown that the unique convergence point is
\begin{equation}
    w^* = e^{d + 1}_1 + \frac{v^*}{\norm{v^*}^2},
\end{equation}
where we picked the right inverse $Y^{-1} = U D^{-1} V^\top$, where $D^{-1}$ contains the inverse of the singular values in the diagonal in the upper part, and is $0$ in the last $d - n$ rows. This is because all $v \in \text{Ker}(Z)$, and if we assumed $v \neq v^*$ to solve Equation \ref{eq:v}, we would have got a contradiction checking the orthogonality constraint over both $v$ and $v^*$.

At this point, after training, an attacker can recover the value of $\norm{v^*}$ feeding the model the synthetic query $z = e^{d + 1}_1$. After this, the attacker can easily obtain $x_i / \norm{v}^2$ feeding the model the query $\tilde z_i = [0, y_i]$, completing the recovery.

If we assume the $z_i$ to be sampled independently from a standard Gaussian distribution, $z_i \sim \mathcal N (0, I_{d + 1})$, we have control on the singular values of $Y$ (in the limit of large $d$), and we can provide more quantitative intuition on the result. If $d \simeq n$, $\norm{v^*}$ becomes very large, and the solution doesn't separate from $e^{d+1}_1$ that much. If $d \gg n$, $\norm{v^*} \simeq 1$. This generates a solution that is almost $0$ in the first component, and larger on the other entries (overfitting with no learning).

\begin{figure*}[ht]
\vskip 0.2in
\begin{center}
\centerline{\includegraphics[width=\textwidth]{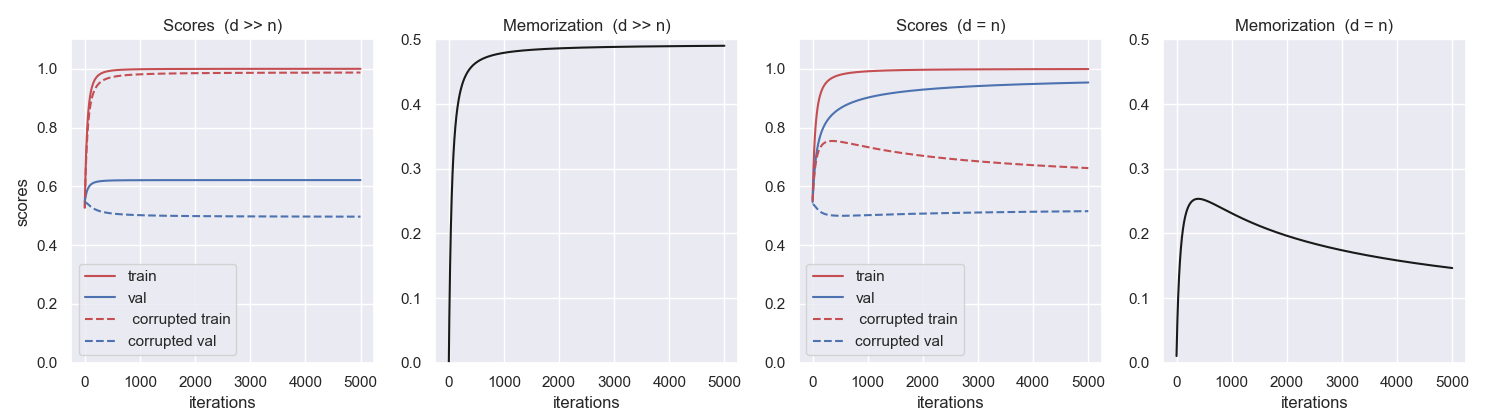}}
\caption{Numerical simulation of the task described in \S\ref{subsec:linear}. Data are sampled from a Gaussian distribution. Scores are computed as $1 - L$, where $L$ is the empirical Loss in Equation \ref{eq:loss}. Memorization is computed as the expected error in reconstructing the $x_i$-s. On the left, simulated dynamics with $n = 300$, $d = 2000$. On the right, $n = d = 1000$.}
\label{fig:synthetic_app}
\end{center}
\vskip -0.2in
\end{figure*}

The dynamics, for finite $t$, is regulated by the spectrum of $D^2 + V^\top X X^\top V$. If $d \gg n$, the second term gets negligible, implying an homogeneous evolution of $w$, which means that Memorization increases with the same rate as the learning. On the other hand, if $d \simeq n$, the dynamics splits, and learning is slower than memorization. This implies that the Relational Memorization, after a critical time, starts to decrease. Numerical simulation of these behaviors, is illustrated in Figure \ref{fig:synthetic_app}.

\subsection{Measure of Relational Memorization}\label{app:4}

We can write:
\begin{equation}
    I(W, Y ; X)
    = \mathbb E_{w \sim p(w|x,y)} \mathbb E_{x, y \sim P_{XY}} \log \frac{p(x | y, w)}{p(x)} 
    \simeq \frac{1}{N} \sum_i \log \frac{p(x_i | y_i, w)}{p(x_i)}.
\end{equation}
In the last approximation, we average over our samples, and we do not average on different training processes. Using the assumption on the optimality of the attack,
\begin{equation}
    I(W, Y ; X) \simeq \frac{1}{N} \sum_i \log \frac{p(x_i | y_i, w)}{p(x_i)} \simeq \frac{1}{N} \sum_i \log \frac{q_w(-, y_i)_{x_i}}{p(x_i)}.
\end{equation}
Defining now $r_{tr}$ the fraction of samples on the training set where $q_w(-, y_i)_{x_i} \simeq 1$, and using $p(x_i) \simeq c$, we get
\begin{equation}
    I(W, Y ; X) \simeq \frac{1}{N} \sum_i \log \frac{q_w(-, y_i)_{x_i}}{p(x_i)} \simeq r_{tr} (- \log c) = H(X)r_{tr} .
\end{equation}

Since the samples are iid, we can write
\begin{equation}
    I(Y_i ; X_i) = I(Y ; X),
\end{equation}
In particular, the previous equation holds also for validation samples. We will indicate them as $(\tilde X, \tilde Y)$. Since they are (jointly) independent on the trained model $W$, we can write 
\begin{equation}
    I(X ; Y) = I(\tilde X ; \tilde Y) = I(W, \tilde Y ; \tilde X).
\end{equation}
This form is exactly the previous computation, but on the validation set.
\begin{equation}
    I(X ; Y) \simeq H(X) r_{val}.
\end{equation}

This gives
\begin{equation}
    \text{mem} = H(X) (r_{tr} - r_{val}).
\end{equation}

To obtain an estimate on Relational Memorization, we need to control $\text{mem}^m = I(W; X) = \sup_\mathcal  R I(\mathcal R(W); X_i)$, where $\mathcal R$ are again recovery attacks. Using the last assumptions, this computation reduces to computing $I(R; X)$, where $R$ is a random variable defined as follows
\begin{equation}
R = 
    \begin{cases} 
      X, \; \text{ with probability } 1/n ;  \\
      \text{i.i.d. to } X \; \text{ with probability } (n-1)/n.
    \end{cases}
\end{equation}
Using the definition of mutual information, and the last assumption we get
\begin{equation}
    \begin{aligned}
    I(R; X) &= \sum p(r_i, x_j) \log \frac{p(r_i, x_j)}{p(r_i)p(x_j)} \\
    &= \sum_{i \neq j} \frac{n-1}{n} c^2 \log \frac{n -1}{n} + 
    \sum_{i = j} \frac{n-1+1/c}{n} c^2 \log \frac{n - 1 +1/c}{n} \\
    &= (1-c) \frac{n-1}{n}  \log \frac{n -1}{n} + 
    c \frac{n-1+1/c}{n} \log \frac{n - 1 +1/c}{n} \\ 
    &\simeq - \frac{1-c}{n} + \frac{1}{n} = \frac{c}{n} = \frac{e^{-H(X)}}{n} \ll H(X).
    \end{aligned}
\end{equation}

Therefore, in these settings, we have
\begin{equation}
    \text{mem}^r = \text{mem} - \text{mem}^m \simeq \text{mem} \simeq H(X) (r_{tr} - r_{val}).
\end{equation}

\end{document}